\documentclass[10pt,journal,compsoc]{IEEEtran}
\usepackage{array}
\usepackage{algorithm}
\usepackage{algpseudocode}
\usepackage{float}
\usepackage{graphicx}
\usepackage{amsmath}
\usepackage{amsthm}
\usepackage{multirow}
\usepackage{subcaption}
\usepackage{subcaption}
\usepackage{tikz}

\ifCLASSOPTIONcompsoc
  \usepackage[nocompress]{cite}
\else
  \usepackage{cite}
\fi
\ifCLASSINFOpdf
\else
\fi
\begin{document}
%
\title{Hamming Encoder: Mining Discriminative $k$-mers for Discrete Sequence Classification}
%
%
%
%

\author{Junjie~Dong,
        Mudi~Jiang,
        Lianyu~Hu,
        Zengyou~He
\IEEEcompsocitemizethanks{\IEEEcompsocthanksitem J. Dong, M. Jiang, L. Hu, Z. He are with School of Software, Dalian University of Technology, Dalian, China.
\protect\\
E-mail: zyhe@dlut.edu.cn }
\thanks{Manuscript received XXXX XX, 2023; revised XXXX XX, 2023.}}

%
%

\markboth{Journal of \LaTeX\ Class Files,~Vol.~xx, No.~xx, March~2023}%
{Shell \MakeLowercase{\textit{et al.}}: Bare Demo of IEEEtran.cls for Computer Society Journals}
%



\IEEEtitleabstractindextext{%
\begin{abstract}
Sequence classification has numerous applications in various fields. Despite extensive studies in the last decades, many challenges still exist, particularly in pattern-based methods. Existing pattern-based methods measure the discriminative power of each feature individually during the mining process, leading to the result of missing some combinations of features with discriminative power. Furthermore, it is difficult to ensure the overall discriminative performance after converting sequences into feature vectors. To address these challenges, we propose a novel approach called Hamming Encoder, which utilizes a binarized 1D-convolutional neural network (1DCNN) architecture to mine discriminative $k$-mer sets. In particular, we adopt a Hamming distance-based similarity measure to ensure consistency in the feature mining and classification procedure. Our method involves training an interpretable CNN encoder for sequential data and performing a gradient-based search for discriminative $k$-mer combinations. Experiments show that the Hamming Encoder method proposed in this paper outperforms existing state-of-the-art methods in terms of classification accuracy.

\end{abstract}

}

\maketitle

\IEEEdisplaynontitleabstractindextext

%
\IEEEpeerreviewmaketitle

\IEEEraisesectionheading{\section{Introduction}\label{sec:introduction}}

%
%
%
%
A discrete sequence is an ordered list of symbols, where symbols represent abstractions of concrete objects. The analysis of such types of sequences is becoming increasingly important as it plays a significant role in various domains, including bioinformatics \cite{zeng2022alphapeptdeep}, text analysis \cite{lodhi2002text}, and financial data analysis \cite{gupta2020comprehensive}.

Classification is one of the most challenging tasks in sequence analysis since there are no explicit features and the dimensionality of the latent feature space is very high. During the past decades, various sequence classification algorithms have been proposed, which can be roughly categorized into three types \cite{xing_brief_2010}: feature-based methods, distance-based methods, and model-based methods. This paper primarily focuses on the feature-based approach, which first converts the sequences into numerical vectors and then employs the standard classification algorithms such as support vector machine (SVM) and neural networks.

The transformation of a sequence into a numerical vector is a complex task, owing to the arbitrary length of sequences. The current popular feature-based approaches \cite{he2019significance,de_smedt_mining_2020,zhou_pattern_2016,egho_user_2017,nguyen2019sqn2vec} is to mine the patterns with discriminative power from the sequences and then transform the sequences into feature vectors. There are several advantages of applying the feature-based approach in practice. Firstly, numerous classifiers for vector data have been proposed and users can select an appropriate classifier based on the specific characteristics of the task. The second point is that this feature-based approach can provide an additional level of interpretability that is important for certain tasks such as gene fragment analysis. As a last point, the extraction of features from sequences has been extensively studied in different fields, providing many choices for effective sequence feature generation.

However, current discriminative pattern mining algorithms for sequence classification have some limitations. Firstly, many existing pattern mining algorithms evaluate each pattern individually, which may miss some combinations of patterns that have significant classification power, e.g. pattern $A$ and pattern $B$ are not discriminative on their own, but their combination may exhibit significant classification power. Secondly, the information behind patterns may not be fully utilized in the feature vector transformation process. This is mainly because the feature vector is typically generated based on whether each pattern is contained by the corresponding sequence. Finally, existing methods may report many redundant patterns, leading to the generation of many redundant features.

Motivated by the above observations, this study proposes a novel approach to detect discriminative patterns for sequence classification. This new method is named Hamming Encoder, which leverages a Binarized Convolutional Neural Network (BCNN) \cite{rastegari2016xnor} architecture and a novel similarity function ($k$-mer Hamming similarity) to transform sequences into feature vectors. The key idea of our method is to train an interpretable neural encoder for sequence data. Our algorithm works as follows: Initially, the sequences are transformed into one-hot vectors, allowing the Hamming Encoder to perform a gradient-based search for discriminative $k$-mers combinations using customized loss functions, such as the Cross-Entropy and supervised contrast losses. Subsequently, the weight of the encoder is then extracted to generate the discriminative $k$-mer set. Finally, the feature vectors are constructed using the extracted $k$-mers and then fed into the classifiers for analysis.

The main contributions of this paper can be summarized as follows:
\begin{itemize}
    \item To identify discriminative pattern combinations in the mining process, Hamming Encoder adopts a gradient-based search, thus it is liable to find a more discriminative pattern set for classification.

    \item In order to better utilize the mined patterns and relieve the redundancy issue, this paper proposes a new similarity function that is compatible with the main framework of Hamming Encoder. Experimental results indicate that our approach outperforms the conventional presence-weighted feature vector representation approach.

    \item This paper proposes an interpretable and flexible neural network-based approach for discriminative pattern mining. Experimental results suggest its efficacy in alleviating overfitting when dealing with data sets of smaller sample size. The proposed Hamming Encoder module can serve as an alternative to conventional neural network modules to enhance interpretability and relief overfitting.

\end{itemize}

The structure of the paper is organized as follows: Section \ref{relativework} introduces related feature-based sequence classification methods and binarized neural networks. In Section \ref{methdo}, we describe our Hamming Encoder method and introduce a new similarity measure. Section \ref{experimentres} presents the experimental results and Section \ref{conclusion} concludes the paper.

\section{Related Work}
\label{relativework}
In this section, we review previous research efforts that are most closely related to our approach. Section \ref{feature-based} provides a brief overview of feature-based methods for sequence classification, which is the primary focus of this paper. In Section \ref{bnn}, we discuss binarized neural networks, which are specially employed to handle $k$-mer strings in our methods.
Section \ref{BNN Mining} introduces the recent progress of BNN-based pattern mining methods.

\subsection{Feature/Pattern-Based Methods}
\label{feature-based}
The most popular feature-based method involves representing a sequence as a vector of presence and absence of $k$-mers (or $n$-grams), where $k$-mer is a small continuous segment of length $k$. However, as $k$ increases, the number of $k$-mers grows exponentially, requiring the use of feature selection  methods to screen out some irrelevant ones \cite{chuzhanova1998feature}.

The concept of subsequence patterns was introduced in \cite{lesh1999mining}. This method views subsequences as pattern representations and provides several standard metrics for filtering patterns. Subsequent studies aimed to reduce redundancy and improve the accuracy of classifiers by applying constraints to refine the set of mined patterns \cite{ji2007mining,pei_constraint-based_2007,srikant1996,zhou_pattern_2016,EXARCHOS2008467,he2019significance,lo2009classification}. Nevertheless, these methods evaluate each pattern individually and potentially miss some discriminative combinations of patterns. Additionally, various similarity functions have been proposed to calculate the distance between patterns and sequences for improving pattern utilization \cite{zhou_pattern_2016,ntagiou2017protein,tsai2015pso,lesh1999mining}. However, these methods do not guarantee the discriminative power of the mined pattern set after converting sequences into feature vectors.


To address the above issue, some studies have combined pattern mining with classifier construction \cite{9411677,ifrim2011bounded,ifrim2008fast,okanohara2009text}. These approaches identify patterns that are optimal for a specific classifier, which may fail to report discriminative pattern combinations as well.

\subsection{Binarized Neural Network}
\label{bnn}
Deep neural networks (DNN) have shown powerful performance in many fields. However, as DNNs become larger, they typically require significant storage and high computational resources. To reduce model parameters and speed up inference for low resources and embedded devices, numerous parameter quantization methods have been proposed \cite{gong2014compressing,wu2016quantized,chen2019metaquant,nagel2022overcoming,lee2021network}. Among them, the binarization technique has undergone significant evolution \cite{rastegari2016xnor,zhou_dorefa-net_2018,bulat_bats_2020,bulat_xnor-net_2019,avidan_adabin_2022}. 
Binarization is a 1-bit quantization that limits data to only two possible values: $-1(0)$ or $+1$. After quantizing the weights of the network, the memory and computational requirements can be significantly reduced without sacrificing much performance. 

However, backpropagation (BP) and stochastic gradient descent (SGD) are not applicable as the binarization function is not differentiable and its derivative value vanishes. Thus, a heuristic method named straight-through estimator (STE) is presented to solve the problem of estimating the gradient \cite{bengio2013estimating}. By bypassing the gradient of the layer in question, the gradient of the binary weights is simply passed through to the real-valued weights. STE makes it possible to apply an optimization strategy like SGD or Adam to update the real-valued weights. Various types of gradient estimators have been proposed according to the characteristics of the task.

While providing a discrete connection between weights and input, these network structures do not intuitively learn interpretable structures related to the input. This paper adopts the same strategy of binarized neural networks to train an interpretable CNN-based autoencoder.

\subsection{Pattern Mining Based on BNN}
\label{BNN Mining}
In recent years, the use of neural networks for mining different types of patterns and rules has received much attention. Many of these approaches leverage binarized neural networks (BNN) to obtain an interpretable structure. This section will focus on the binarization-based approach, which is closely relative to our method.

In \cite{Wang2019TransparentCW,qiao2020learning,wang2021scalable,wang2022tabular}, binarized neural networks are utilized for mining classification rules from non-sequential data. Concept Rule Sets (CRS) \cite{Wang2019TransparentCW} uses gradient descent and random binarization to learn rule sets efficiently, but it is limited by its logical activation functions and training method when applied to large datasets. To address this issue, Rule-based Representation Learner (RRL) \cite{wang2021scalable} was introduced. NR-Net \cite{qiao2020learning} formulates the task of learning accurate and interpretable decision rule sets as a neural network training problem, and presents a two-layer neural network architecture that can be directly mapped to a set of interpretable decision rules. To further improve performance, thresholds are added to CT-Net for tabular classification \cite{wang2022tabular}.

Several methods leveraging neural networks have also been proposed for pattern set mining \cite{fischer_differentiable_2021,itkar2014efficient,dong2005distributed,kamruzzaman2011new,baez2006identification}. In particular,  Binaps \cite{fischer_differentiable_2021} uses constrained networks with binarized weights to mine patterns on large-scale datasets.

The methods described in the previous paragraphs are not tailored for sequential data. Recently, there have been some methods proposed for applying neural networks to sequential pattern mining \cite{jamshed2020deep,9207461,Jiang2021PredictiveSP}. However, these methods are not designed to tackle sequence classification problems.

In the domain of sequence classification, CR2N \cite{colleryneural} combines a 1D-convolutional neural network (1DCNN) and a base rule model architecture for discovering classification rules with local or global patterns. The method proposed in this paper is structurally similar to CR2N to some extent, as both use the 1DCNN architecture. The main difference is that the global pattern mined by CR2N is the disjunction rule, whereas the mined pattern in our method is a set of discriminative substrings. To achieve this target, we employ a special binarization strategy for 1DCNNs. This strategy allows the weights of each kernel to be updated during training and can be interpretably translated into a segment of $k$-mer.

\begin{figure*}[!htbp]%
    \centering
    {\includegraphics[scale = 1]{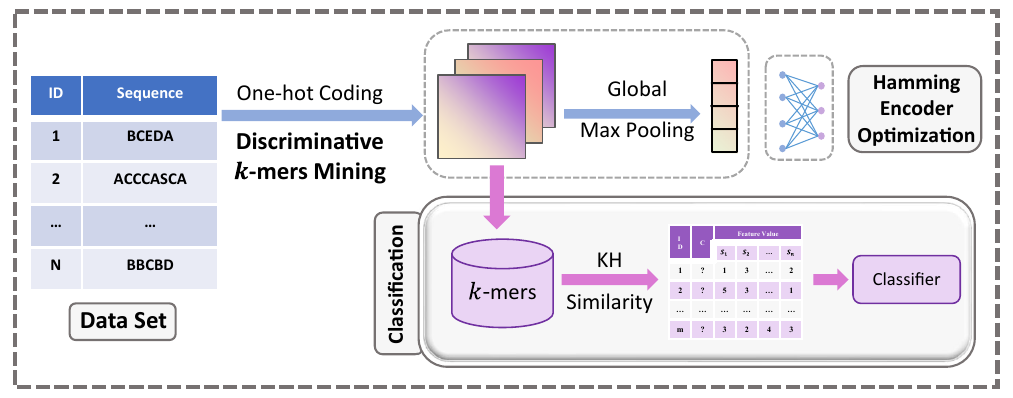}}
    \caption{The workflow of Hamming Encoder involves converting sequence data into one-hot matrices, optimizing the Hamming Encoder, and translating its weights into a discriminative $k$-mer set. The feature vector is generated using the KH similarity measure with the $k$-mers, and a classifier is then adopted for classification.} \label{liuchengtu}
\end{figure*}

\section{Method}
\label{methdo}
In this section, we first present notations in Section \ref{notation}  and then provide a description of previous work in Section \ref{Previous Works}. Section \ref{overview} then gives an overview of the method. Subsequently, the binarizing process of Hamming Encoder is then introduced in Section \ref{Hamming Encoder}. In Section \ref{khsimilarity}, we show the consistency of Hamming distance-based similarity and the global max pooling operation.

\subsection{Notation}
Considering an item set $ I = \left\{i_1 , i_2 , \dots , i_m\right\}$ with $m$ distinct items. A sequence $s =\langle s_1 , s_2 , \dots , s_l \rangle$ is an ordered list over $I$ if each $s_i \in I$. For a labeled sequence data set $D$, each record is composed of a sequence $s$ and a class label $c_i \in C$, where $C = \left\{c_1, c_2, \dots, c_d\right\}$ is a set of $d$ distinct class labels. The main idea of this work is to mine a set of discriminative $k$-mers over data set $D$ with an interpretable CNN encoder, where the weight matrix is denoted as $\mathbf{w}$ and $\mathbf{w}_b$ for its binarized version. The one-hot sequence matrices are defined as $\mathbf{a}$. The Hamming Encoder encoder layer is then defined as $f_{\mathbf{w_b}}(\mathbf{a})$
\label{notation}. Partial derivatives of a function $f$ with respect to a variable $x$ are denoted as $\frac{\partial f}{\partial x}$.

\subsection{Previous Works}
\label{Previous Works}
For full-precision one dimensional convolutional neural networks (1DCNN), given a  weight matrix $\mathbf{w} $, and a one-hot sequence matrix $\mathbf{a}$:
\begin{equation}
    \mathbf{a}_{i, j}=\left\{
    \begin{array}{ll}
        1, & \text { if } s_{j} \text { is the \textit{i}-th item of } I, \\
        0, & \text { Otherwise. }
    \end{array}\right.
\end{equation} 
The output $\mathbf{z} $ can be obtained by the convolution operation in Eq. \ref{cnn_full}:
\begin{equation}
    \label{cnn_full}
    \mathbf{z}=\sigma(\mathbf{w} \otimes \mathbf{a}),
\end{equation}
where $\sigma(\cdot)$ is a non-linear function and $\otimes$ represents the convolutional operation. For binarization, existing BNNs generally apply a $sign$ function:
\begin{equation}
    sign(x)=\left\{\begin{array}{ll}
        +1, & \text { if } x \geq 0 \\
        -1, & \text { otherwise. }
    \end{array}\right.
\end{equation}
With quantization, the output $\mathbf{z}$ then is:
\begin{equation}
    \mathbf{z} = \sigma (  sign(\mathbf{w}) \otimes  sign( \mathbf{a} )).
\end{equation}
However, such a binarization strategy does not generate interpretable structures and the result is just a pile of matrices of $-1$ and $+1$.

\subsection{Overview of the Method}
\label{overview}
For a given sequence database, the purpose of this paper is to find a set of discriminative $k$-mers for sequence classification. In particular, we do not want to miss some pattern combinations with high discriminability. In neural network structure, the encoder is commonly used for extracting features for downstream tasks, and the CNN-based encoder has shown its powerful feature extraction capability for sequence classification. However, the connections between inputs and neurons in neural networks are often non-symbolic and non-linear, making them inherently difficult for humans to comprehend.

Here, we propose the Hamming Encoder, where weights are binarized with 0 and 1 during the forward pass. It is an interpretable CNN encoder where each kernel can be translated into a $k$-mer. Therefore, mining discriminative patterns is formulated in this paper as an optimization problem for the CNN encoder. The workflow of the Hamming Encoder is summarized in Fig. \ref{liuchengtu}. Each sequence in the data set is first converted into a $|I| \times |s|_{max}$ one-hot matrix, where $|I|$ is the size of the item set, and $|s|_{max}$ denotes the maximal sequence length in the data set. In order to obtain a powerful Hamming Encoder, a fully connected layer was added, as it represents feature information after a mutual combination. After optimizing the Hamming Encoder with a loss function, its weight is then transformed into a $k$-mer set. Subsequently, we propose a new method called KH similarity, which calculates the similarity between sequences and discriminative $k$-mers to generate feature vectors. KH similarity is equivalent to the global max pooling operation used in the network structure, which enhances the discriminative power of the mined $k$-mers by converting them into feature vectors with more information.

\subsection{Hamming Encoder}
\label{Hamming Encoder}
The structure of Hamming Encoder is a special binarized CNN layer followed by a full-precision FNN layer. Each kernel (or channel) of the binarized CNN layer has binary weights $\mathbf{w}_b$ for the forward process and a continuous version of weight $\mathbf{w}$ for backpropagation, where the binary weights $\mathbf{w}_b$ can be translated into $k$-mers. By optimizing the encoder, the discriminative $k$-mer set can be obtained. Algorithm \ref{alg:Hamming Encoder} presents the details of the training process of Hamming Encoder. The following sections describe how to binarize in the forward pass and backward pass process.
\begin{algorithm}[t]
    \caption{Hamming Encoder Mining}
    \label{alg:Hamming Encoder}
    \begin{algorithmic}[0]
        \State \textbf{Input:} A sequence database $D$, initial weight $\mathbf{w}^1$, number of epochs $e_{max}$, batchsize $N$, itemset $I$.
        \State \textbf{Output:} Discriminative $k$-mer set $P$.
    \end{algorithmic}
    \begin{algorithmic}[1]
        \State $\mathbf{t} \gets 1$  \Comment{Step}
        \State $\mathbf{a} \gets \text{one-hot}(I,D)$
        \For {$e=1 \dots e_{max}$} \Comment{Epochs}
        \For {$i=1\dots n$}\Comment{For each batch}

        \State \textbf{//Forward pass}
        \State $\mathbf{w}^t_{bcnn} \gets  \mathbf{Q}( \mathbf{w}^t_{cnn}) $; \Comment{Binarized weights}
        \State $\mathbf{z} \gets \mathbf{w}^t_{bcnn} \otimes \mathbf{a}_i$; \Comment{Convolution}
        \State $\mathbf{k} \gets \text{GlobalMaxPooling}(\mathbf{z})$;
        \State $\mathbf{y} \gets  \mathbf{k}\mathbf{w}_{fnn}^T$ \Comment{FNN layer}
        \State $\mathcal{L} \gets \text{loss} ( D[i]; \mathbf{y} )$; \Comment{Compute loss}
        \State \textbf{//Backward pass}
        \State $g_{\mathbf{w}_{cnn}} \gets \text{backward} (\mathcal{L})$;
        \State $\mathbf{w}^{t+1}_{cnn} \gets \text{update} 
        (\mathbf{w}^t_{cnn}, g_{\mathbf{w}_{cnn}})$; \Comment{Optimize}
        \EndFor
        \EndFor

        \State $P \gets \text{inverse one-hot}(I, \mathbf{w}_{bcnn})$
        \Comment{Extract $k$-mers}
        \State \textbf{Output:} $P$
    \end{algorithmic}
\end{algorithm}

\subsubsection{The forward pass}
In our method, the concrete values in the weight matrix are not considered. Instead, for a given weight matrix $\mathbf{w}$, we create a binarized matrix $\mathbf{w}_{b}$ as Eq. \ref{hammingbnn}, such that the elements of $\mathbf{w}_{b}$ are set to $1$ at the corresponding position of the maximum value in each column of the original matrix $\mathbf{w}$, and $0$ otherwise.
\begin{equation}
    \label{hammingbnn}
    \mathbf{Q}(\mathbf{w})_{i,j}=\left\{\begin{array}{ll}
        1, & \text{if } i=\arg\max_{k}\mathbf{w}_{k,j} \\
        0, & \text{otherwise.}
    \end{array}\right.
\end{equation}
Thus, with the one-hot sequence matrix $\mathbf{a} \in \mathcal{R}^{|\mathcal{I}| \times|s| _{max} }$, the output of binarized CNN layer is:
\begin{equation}
    f_{\mathbf{w_b}}(\mathbf{a}) =  \mathbf{w_b} \otimes \mathbf{a}.
\end{equation}
Subsequently, a global max pooling operation is applied to the output of the binarized CNN layer to obtain the feature vector $\mathbf{k}$. We will show the vector is equivalent to the feature value according to the KH similarity defined in Section \ref{sec:khsimilarity}.

\begin{equation}
\label{maxpoolingeq}
    \mathbf{k}_i = \max_{j}  (f_{\mathbf{w_b}}(\mathbf{a})_{i,j}).
\end{equation}

The feature vector $\mathbf{k}$ is then fed into a fully connected neural network (FNN) to obtain the final output $\mathbf{y}$:
\begin{equation}
    \mathbf{y} = \mathbf{k} \mathbf{w}_{fnn}^T.
\end{equation}

\subsubsection{The backward pass}
For mining the discriminative $k$-mers, we minimize the loss function $\mathcal{L}$. By taking the output $\mathbf{y}$ and the ground truth $\hat{\mathbf{y}} = D[i,j]$ as input, the loss function of Cross-Entropy is defined as:
\begin{equation}
    \mathcal{L}=-\frac{1}{N} \sum_{i=1}^{N} \sum_{j=1}^{|C|} \mathbf{y}_{i j} \log \left(\hat{\mathbf{y}}_{i j}\right),
\end{equation}
where $N$ is the batch size, $|C|$ is the number of classes. With the loss value across the batch, we can calculate the gradient of the loss function with respect to the weights of the Hamming Encoder layer. For updating the weights, the straight-through-estimator (STE) \cite{bengio2013estimating} is employed in the Hamming Encoder layer to estimate the backward gradients for quantization. STE assigns the incoming gradients to a threshold operation to the outcoming gradients:
\begin{equation}
    \frac{\partial \mathcal{L}}{\partial \mathbf{w}}=\frac{\partial \mathcal{L}}{\partial \mathbf{w}_{b}}.
\end{equation}
The STE operation is applied to the weight matrix $\mathbf{w}$ in the Hamming Encoder layer, and the gradient of the loss function with respect to the weights of the Hamming Encoder layer is calculated as:
\begin{equation}
    \frac{\partial \mathcal{L}}{\partial \mathbf{w}} = \frac{\partial \mathcal{L}}{\partial \mathbf{y}} \cdot \frac{\partial \mathbf{y}}{\partial \mathbf{k}} \cdot \frac{\partial \mathbf{k}}{\partial  f_{\mathbf{w_b}}(\mathbf{a})} \cdot \frac{\partial  f_{\mathbf{w_b}}(\mathbf{a})}{\partial \mathbf{w}_{b}}.
\end{equation}

\begin{figure*}[!h]%
    \centering
    {\includegraphics[scale = 0.9]{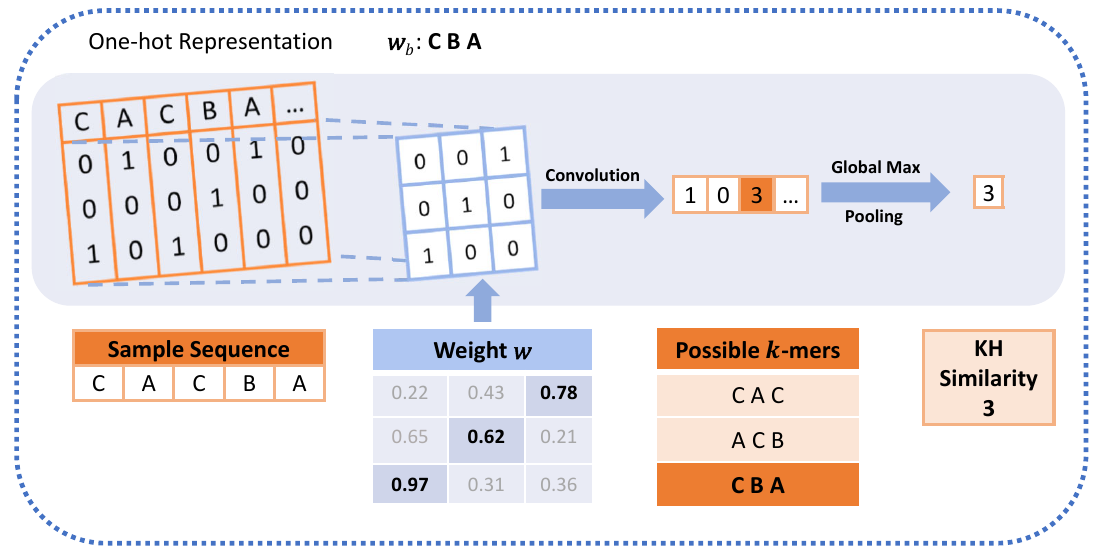}}
    \caption{An example for illustrating the consistency between the KH similarity and the global max pooling operation in neural networks.} \label{KH similarity}
\end{figure*}
\subsubsection{Discriminative $k$-mer set Mining}
For an optimized Hamming Encoder, we first read the weights $\mathbf{w}_{cnn}$ from the saved network parameters and then apply the binarization strategy $\mathbf{Q}(\cdot)$ to it. This allows us to obtain the binarized weights computed during the forward pass, denoted as $\mathbf{w}_{bcnn}$. Subsequently, we transform the weights of the matrix into $k$-mers based on the item set $I$, following the same order used to transform the input sequence into a one-hot matrix. Thus, for each $\mathbf{w}_{bcnn}$, we can identify a corresponding $k$-mer $p$:

\begin{equation}
\label{weighttop}
p_n = i_j \in I \quad \text{for } j = 1,2,\dots,m \text{ s.t. } \mathbf{w}_{j,n}=1,
\end{equation}
where $I$ denotes the set of elements, $p_n$ denotes the $n$-th element in the translated sequence, $i_j$ denotes the $j$-th element in the set of elements, and $\mathbf{w}_{j,n}$ denotes the element in the $j$-th row and $n$-th column of the weight matrix $\mathbf{w}_{bcnn}$. The formula means that for each $n$ column of the matrix $\mathbf{w}_{bcnn}$, find the row $j$ with value $1$ and add the element $i_j$ corresponding to that row to the $n$-th position of the $k$-mer $p$. 

Here is an example of the conversion of the weight matrix $\mathbf{w}_{cnn}$ into a $k$-mer $p$.
\begin{equation*}
\mathbf{w}_{cnn} = \begin{pmatrix}
    0.22 & 0.43 & 0.78\\
    0.65 & 0.62 & 0.21\\
    0.97 & 0.31 & 0.36
    \end{pmatrix},
\mathbf{w}_{bcnn} =
\begin{pmatrix}
    0 & 0 & 1\\
    0 & 1 & 0\\
    1 & 0 & 0
    \end{pmatrix}.
\end{equation*}
For a given item set $I = \{A,B,C\}$, the weight matrix $\mathbf{w}_{bcnn}$ is then transformed into $\langle CBA \rangle$.

\subsection{$k$-mer Hamming (KH) Similarity}
\label{sec:khsimilarity}
To ensure consistency in the conversion of $k$-mer into feature vectors and the values obtained in the network after the global max pooling operation, we propose a $k$-mer Hamming (KH) similarity measure. This measure involves computing the Hamming distance between a substring of a sliding window of length $k$ and a $k$-mer $p \in P$, as shown in the following equation.
\label{khsimilarity}

\begin{equation}
    \mathrm{Sim}(p, s) = k - \min_{t\in S_k(s)} \{ {H(t, p)} \},
\end{equation}
where $S_k(s)$ represents the set of all substrings with length $k$ in instance $s$, and  $H(t, p)$ represents the Hamming distance between the $k$-mer sequence $p$ and the substring $t$:

\begin{equation}
    H(t, p) =\sum_{i=1}^k (t[i] \oplus p[i]),
\end{equation}
where $\oplus$ denotes the XOR operation.

\textbf{Theorem 1.} Given a kernel weight $\mathbf{w}_b$ and its corresponding $k$-mer $p$ under one-hot encoding, for any given sequence $\mathbf{s}$ and its one-hot matrix $\mathbf{a}$, the $k$-mer Hamming similarity is equal to the global max pooling on the output of convolutional operation on $a$ with $\mathbf{w}_b$ on channel $j$:

\begin{equation}
   \mathrm{Sim}(p, s) = \max_{j}  (f_{\mathbf{w_b}}(\mathbf{a})_{i,j}).
\end{equation}

\begin{proof}
Assuming $\mathbf{w}_b = \{ w_1,w_2,\dots,w_k \},    
$
where $w_i$ is the $i$-th column vector of $\mathbf{w}_b$ and can be converted into $p_i$ according to Equation \ref{weighttop}. In the same way, $\mathbf{a}$ can also be represented as the form of the combination of column vector $\mathbf{a} =\{ a_1,a_2,\dots,a_{|s|} \},$ then the output of convolution operation $ \mathbf{y}  =  f_{\mathbf{w_b}}(\mathbf{a})$,
\begin{equation}
    \mathbf{y}_i = \sum_{{i=1}}^{|s|- k + 1} \sum_{{j=0}}^{k-1} a_{i+j} \cdot w_i.
\end{equation}

Known that $a_i$ is the one-hot vector, and the weight matrix $\mathbf{w}_b$ can be converted into $p$ using the same encoding relationship. Based on the orthogonality of one-hot encoding, we know that the dot product of two one-hot vectors equals 1 if and only if the two vectors are identical.

That is, we could say the convolutional operation is actually counting the number of identical vectors in the kernel width $k$:
\begin{equation}
    \mathbf{y}_i  = k - H(\mathbf{a}_{i:i+k-1},\mathbf{w}_b).
\end{equation}

Subsequently, the max pooling operation is adopted according to the Equation \ref{maxpoolingeq}:
\begin{equation}
\mathbf{z} = \max_{i \in |s|- k + 1} \{\mathbf{y}_i\} = k - \min_{i \in |s|- k + 1} \{H(\mathbf{a}_{i:i+k-1},\mathbf{w}_b)\}.
\end{equation}
With the sliding window, $\mathbf{a}_{i:i+k-1}$ is then actually can be converted into the $k$-mer $t$ in sequence, and $\mathbf{w}_b$ can also be inversely encoded into $p$. Thus, KH similarity proposed in this paper is equivalent to the global max pooling operation.
\end{proof}

In Fig. \ref{KH similarity}, we illustrate how the KH similarity is computed and its consistency with the global max pooling operation. Specifically, we binarize the weight matrix $\mathbf{w}$ to obtain the $\mathbf{w_b}$ matrix, which is utilized to compute the output $\mathbf{y}$ from a matrix $\mathbf{a}$ that is derived from the one-hot encoding of the sequence $CACBA$. By performing a global max pooling operation on the output $\mathbf{y}$, we then obtain the feature vector $\mathbf{z}$, which is equivalent to the feature value in the KH similarity measure. For this example, where $k=3$, $p = \langle CBA \rangle$, $s = CACBA$, and $S_k(I) = \left\{CAC, ACB, CBA\right\}$, the Hamming distance computation yields a KH similarity value of $3$.

\section{Experimental Results}
\label{experimentres}

In this section, we begin by evaluating the performance of our proposed method on eleven benchmark data sets and compare it with several feature-based sequence classification algorithms. To investigate the distinctions between the Hamming Encoder and the baseline CNN, as well as the widely-used BCNN, a series of experiments were executed as well. The experiments were performed on a PC with NVIDIA 3060 GPU and AMD Ryzen 9 5900X CPU with 32 GB memory. All the reported accuracies are obtained by repeating the 5-fold cross-validation 5 times to record the average result. When evaluating CR2N-based methods for handling data sets with more than 2 classes, we employ the One-vs-One strategy to extend the binary classifier.

\subsection{Data Sets and Setup}
\begin{table}[!htbp]
    \centering
    \caption{Some characteristics of the datasets used in the performance comparison.}
    \label{data set}
    \begin{tabular}{cccccc}
        \hline Datasets & $|D|$ & $|I|$ & min$l$ & max$l$ & $|C|$ \\
        \hline
        Aslbu          & 424   & 250   & 2      & 54     & 7   \\
        Auslan2        & 200   & 16    & 2      & 18     & 10  \\
        Context        & 240   & 94    & 22     & 246    & 5   \\
        Epitope        & 2392  & 20    & 9      & 21     & 2   \\
        Gene           & 2942  & 5     & 41     & 216    & 2   \\
        Pioneer        & 160   & 178   & 4      & 100    & 3   \\
        Question       & 1731  & 3612  & 4      & 29     & 2   \\
        Robot          & 4302  & 95    & 24     & 24     & 2   \\
        Skating        & 530   & 82    & 18     & 240    & 7   \\
        Reuters        & 1010  & 6380  & 4      & 533    & 4   \\
        Unix           & 5472  & 1697  & 1      & 1400   & 4   \\
        \hline
    \end{tabular}

\end{table}

\begin{table*}
\centering
\caption{Performance comparison of Hamming Encoder, MiSeRe, Sqn2Vec, iBCM, SGT, SCIP, CR2N and SeqDT in terms of the classification accuracy.}
\label{tab:results}
\begin{tabular}{cccccccc|cc|ccc} 
\hline
Datasets                  & Classifier & \begin{tabular}[c]{@{}c@{}}\textbf{Hamming}\\\textbf{Encoder}\end{tabular} & MiSeRe         & \begin{tabular}[c]{@{}c@{}}Sqn2Vec\\ SEP\end{tabular} & \begin{tabular}[c]{@{}c@{}}Sqn2Vec\\ SIM\end{tabular} & iBCM       & SGT                  & Classifier & SCIP                      & \begin{tabular}[c]{@{}c@{}}CR2N\\ (local)\end{tabular} & \begin{tabular}[c]{@{}c@{}}CR2N\\ (global)\end{tabular} & SeqDT                            \\ 
\hline
\multirow{4}{*}{aslbu}    & SVM        & \textbf{0.646}                                                                   & 0.571          & 0.498                                                 & 0.633                                                 & 0.557      & 0.427                & SCII\_HAR  & 0.54                      & \multirow{4}{*}{0.438}                               & \multirow{4}{*}{0.344}                                & \multirow{4}{*}{0.565}           \\
                          & NB         & 0.581                                                                   & 0.548          & 0.298                                                 & 0.554                                                 & 0.547      & 0.535                & SCII\_MA   & 0.526                     &                                                      &                                                       &                                  \\
                          & KNN        & 0.646                                                                   & 0.574          & 0.544                                                 & 0.591                                                 & 0.488      & 0.414                & SCIS\_HAR  & 0.553                     &                                                      &                                                       &                                  \\
                          & DT         & 0.581                                                                   & 0.565          & 0.45                                                  & 0.484                                                 & 0.476      & 0.521                & SCIS\_MA   & 0.536                     &                                                      &                                                       &                                  \\ 
\\
\multirow{4}{*}{auslan2}  & SVM        & 0.274                                                                   & 0.3            & 0.29                                                  & 0.31                                                  & 0.285      & 0.26                 & SCII\_HAR  & 0.1                       & \multirow{4}{*}{0.23}                                & \multirow{4}{*}{0.215}                                & \multirow{4}{*}{\textbf{0.327}}  \\
                          & NB         & 0.306                                                                   & 0.304          & 0.26                                                  & 0.29                                                  & 0.29       & 0.291                & SCII\_MA   & 0.095                     &                                                      &                                                       &                                  \\
                          & KNN        & 0.294                                                                   & 0.302          & 0.31                                                  & 0.2                                                   & 0.29       & 0.243                & SCIS\_HAR  & 0.2                       &                                                      &                                                       &                                  \\
                          & DT         & 0.304                                                                   & 0.314          & 0.27                                                  & 0.27                                                  & 0.285      & 0.266                & SCIS\_MA   & 0.175                     &                                                      &                                                       &                                  \\ 
\\
\multirow{4}{*}{context}  & SVM        & \textbf{0.953}                                                          & 0.927          & 0.933                                                 & 0.9                                                   & 0.896      & 0.925                & SCII\_HAR  & 0.613                     & \multirow{4}{*}{0.238}                               & \multirow{4}{*}{0.167}                                & \multirow{4}{*}{0.854}           \\
                          & NB         & 0.836                                                                   & 0.938          & 0.9                                                   & 0.9                                                   & 0.896      & 0.741                & SCII\_MA   & 0.617                     &                                                      &                                                       &                                  \\
                          & KNN        & 0.929                                                                   & 0.927          & 0.933                                                 & 0.9                                                   & 0.871      & 0.719                & SCIS\_HAR  & 0.796                     &                                                      &                                                       &                                  \\
                          & DT         & 0.833                                                                   & 0.841          & 0.6                                                   & 0.542                                                 & 0.813      & 0.832                & SCIS\_MA   & 0.867                     &                                                      &                                                       &                                  \\ 
\\
\multirow{4}{*}{epitope}  & SVM        & 0.922                                                                   & 0.834          & 0.802                                                 & 0.8                                                   & 0.885      & 0.919                & SCII\_HAR  & 0.684                     & \multirow{4}{*}{0.561}                               & \multirow{4}{*}{0.592}                                & \multirow{4}{*}{0.836}           \\
                          & NB         & 0.716                                                                   & 0.588          & 0.779                                                 & 0.761                                                 & 0.698      & 0.734                & SCII\_MA   & 0.712                     &                                                      &                                                       &                                  \\
                          & KNN        & 0.872                                                                   & \textbf{0.924} & 0.863                                                 & 0.841                                                 & 0.768      & 0.822                & SCIS\_HAR  & 0.705                     &                                                      &                                                       &                                  \\
                          & DT         & 0.869                                                                   & 0.842          & 0.813                                                 & 0.8                                                   & 0.879      & 0.858                & SCIS\_MA   & 0.721                     &                                                      &                                                       &                                  \\ 
\\
\multirow{4}{*}{gene}     & SVM        & \textbf{1}                                                              & \textbf{1}     & 0.999                                                 & 0.976                                                 & \textbf{1} & \textbf{1}           & SCII\_HAR  & \textbf{1}                & \multirow{4}{*}{0.815}                               & \multirow{4}{*}{0.798}                                & \multirow{4}{*}{\textbf{1}}      \\
                          & NB         & \textbf{1}                                                              & \textbf{1}     & \textbf{1}                                            & \textbf{1}                                            & \textbf{1} & \textbf{1}           & SCII\_MA   & \textbf{1}                &                                                      &                                                       &                                  \\
                          & KNN        & \textbf{1}                                                              & \textbf{1}     & \textbf{1}                                            & \textbf{1}                                            & \textbf{1} & \textbf{1}           & SCIS\_HAR  & \textbf{1}                &                                                      &                                                       &                                  \\
                          & DT         & \textbf{1}                                                              & \textbf{1}     & 0.967                                                 & \textbf{1}                                            & \textbf{1} & \textbf{1}           & SCIS\_MA   & \textbf{1}                &                                                      &                                                       &                                  \\ 
\\
\multirow{4}{*}{pioneer}  & SVM        & 0.99                                                                    & 0.989          & 0.788                                                 & 0.825                                                 & 0.95       & 0.884                & SCII\_HAR  & 0.963                     & \multirow{4}{*}{0.788}                               & \multirow{4}{*}{0.663}                                & \multirow{4}{*}{\textbf{1}}      \\
                          & NB         & 0.935                                                                   & 0.891          & 0.975                                                 & 0.963                                                 & 0.963      & 0.888                & SCII\_MA   & 0.963                     &                                                      &                                                       &                                  \\
                          & KNN        & 0.994                                                                   & 0.975          & 0.975                                                 & 0.988                                                 & 0.775      & 0.624                & SCIS\_HAR  & 0.975                     &                                                      &                                                       &                                  \\
                          & DT         & 0.932                                                                   & 0.988          & 0.788                                                 & 0.825                                                 & 1          & 0.985                & SCIS\_MA   & 0.975                     &                                                      &                                                       &                                  \\ 
\\
\multirow{4}{*}{question} & SVM        & \textbf{0.956}                                                          & 0.902          & 0.789                                                 & 0.881                                                 & 0.879      & /                    & SCII\_HAR  & 0.833                     & \multirow{4}{*}{0.711}                               & \multirow{4}{*}{0.821}                                & \multirow{4}{*}{0.942}           \\
                          & NB         & 0.925                                                                   & 0.87           & 0.736                                                 & 0.762                                                 & 0.837      & /                    & SCII\_MA   & 0.785                     &                                                      &                                                       &                                  \\
                          & KNN        & 0.918                                                                   & 0.897          & 0.828                                                 & 0.874                                                 & 0.881      & /                    & SCIS\_HAR  & 0.846                     &                                                      &                                                       &                                  \\
                          & DT         & 0.944                                                                   & 0.885          & 0.717                                                 & 0.809                                                 & 0.884      & /                    & SCIS\_MA   & 0.837                     &                                                      &                                                       &                                  \\ 
\\
\multirow{4}{*}{robot}    & SVM        & 0.954                                                                   & 0.913          & 0.84                                                  & 0.834                                                 & 0.855      & 0.933                & SCII\_HAR  & 0.795                     & \multirow{4}{*}{\textbf{0.971}}                      & \multirow{4}{*}{0.806}                                & \multirow{4}{*}{0.885}           \\
                          & NB         & 0.781                                                                   & 0.826          & 0.808                                                 & 0.822                                                 & 0.795      & 0.855                & SCII\_MA   & 0.822                     &                                                      &                                                       &                                  \\
                          & KNN        & 0.936                                                                   & 0.937          & 0.945                                                 & 0.949                                                 & 0.818      & 0.894                & SCIS\_HAR  & 0.817                     &                                                      &                                                       &                                  \\
                          & DT         & 0.893                                                                   & 0.9            & 0.811                                                 & 0.778                                                 & 0.792      & 0.899                & SCIS\_MA   & 0.819                     &                                                      &                                                       &                                  \\ 
\\
\multirow{4}{*}{skating}  & SVM        & 0.352                                                                   & 0.293          & \textbf{0.37}                                         & 0.321                                                 & 0.257      & 0.31                 & SCII\_HAR  & 0.181                     & \multirow{4}{*}{0.179}                               & \multirow{4}{*}{0.145}                                & \multirow{4}{*}{0.276}           \\
                          & NB         & 0.275                                                                   & 0.29           & 0.336                                                 & 0.321                                                 & 0.225      & 0.271                & SCII\_MA   & 0.181                     &                                                      &                                                       &                                  \\
                          & KNN        & 0.312                                                                   & 0.241          & 0.302                                                 & 0.34                                                  & 0.194      & 0.229                & SCIS\_HAR  & 0.189                     &                                                      &                                                       &                                  \\
                          & DT         & 0.275                                                                   & 0.258          & 0.226                                                 & 0.23                                                  & 0.223      & 0.256                & SCIS\_MA   & 0.191                     &                                                      &                                                       &                                  \\ 
\\
\multirow{4}{*}{reuters}  & SVM        & 0.975                                                                   & 0.962          & \textbf{0.984}                                        & 0.974                                                 & 0.947      & /                    & SCII\_HAR  & 0.951                     & \multirow{4}{*}{0.536}                               & \multirow{4}{*}{0.597}                                & \multirow{4}{*}{0.892}           \\
                          & NB         & 0.929                                                                   & 0.903          & 0.921                                                 & 0.905                                                 & 0.85       & /                    & SCII\_MA   & 0.953                     &                                                      &                                                       &                                  \\
                          & KNN        & 0.952                                                                   & 0.899          & 0.96                                                  & 0.96                                                  & 0.824      & /                    & SCIS\_HAR  & 0.957                     &                                                      &                                                       &                                  \\
                          & DT         & 0.904                                                                   & 0.912          & 0.826                                                 & 0.741                                                 & 0.892      & /                    & SCIS\_MA   & 0.956                     &                                                      &                                                       &                                  \\ 
\\
\multirow{4}{*}{unix}     & SVM        & 0.89                                                                    & \textbf{0.915} & 0.899                                                 & 0.872                                                 & 0.88       & /                    & SCII\_HAR  & 0.837                     & \multirow{4}{*}{0.52}                                & \multirow{4}{*}{0.598}                                & \multirow{4}{*}{0.909}           \\
                          & NB         & 0.831                                                                   & 0.768          & 0.703                                                 & 0.566                                                 & 0.834      & /                    & SCII\_MA   & 0.838                     &                                                      &                                                       &                                  \\
                          & KNN        & 0.86                                                                    & 0.873          & 0.892                                                 & 0.891                                                 & 0.866      & /                    & SCIS\_HAR  & 0.857                     &                                                      &                                                       &                                  \\
                          & DT         & 0.878                                                                   & 0.899          & 0.776                                                 & 0.756                                                 & 0.874      & /                    & SCIS\_MA   & 0.842                     &                                                      &                                                       &                                  \\ 
\hline
\multicolumn{2}{c}{Average Acc }       & \textbf{0.776}                                                                   & 0.761          & 0.721                                                 & 0.726                                                 & 0.732      & \multicolumn{1}{c}{} &            & \multicolumn{1}{c}{0.7}   & 0.544                                                & 0.522                                                 & 0.771                            \\
\hline
\multicolumn{2}{c}{Average Rank}       & \textbf{2.182}                                                                   & 3.364          & 5.183                                                 & 5.182                                                 & 4.818      & \multicolumn{1}{c}{} &            & \multicolumn{1}{c}{5.455} & 7.636                                                & 8.454                                                 & 2.727                            \\
\hline
\end{tabular}
\end{table*}

\begin{table*}
\centering
\caption{Performance comparison of all feature-based methods and Hamming Encoder in terms of the average accuracy on different classifiers.}
\label{tab:averageacc}
\begin{tabular}{cccccccc} 
\hline
Method & \textbf{Hamming Encoder} & MiSeRe & Sqn2VecSEP & Sqn2VecSIM & iBCM  & Classifier & SCIP   \\ 
\hline
SVM    & \textbf{0.810}      & 0.782  & 0.745      & 0.757      & 0.763 & SCII\_HAR  & 0.681  \\
NB     & \textbf{0.738}      & 0.721  & 0.701      & 0.713      & 0.721 & SCII\_MA   & 0.678  \\
KNN    & \textbf{0.792}      & 0.777  & 0.777      & 0.776      & 0.707 & SCIS\_HAR  & 0.719  \\
DT     & \textbf{0.765 }     & 0.764  & 0.659      & 0.658      & 0.738 & SCIS\_MA   & 0.720  \\
\hline
\end{tabular}

\end{table*}

For evaluating the performance of the proposed method, the following eleven data sets are used: Aslbu \cite{di2013two}, Auslan2 \cite{di2013two}, Context \cite{maggi2012efficient}, Epitope \cite{maggi2016semantical}, Gene \cite{gama2014survey}, Pioneer \cite{di2013two}, Question \cite{lam2014mining}, Robot \cite{zhang2015ccspan}, Skating \cite{di2013two}, Reuters \cite{zhang2015ccspan}, and Unix \cite{zhang2015ccspan}. The details of these data sets are summarized in Table \ref{data set}, where $|D|$ refers to the number of records, $|I|$ is the number of items, min$l$ and max$l$ are the minimum and maximum length of the sequences, respectively, and $|C|$ is the number of classes.

\subsubsection{Baseline Methods}
We evaluate the performance of our proposed method by comparing it with the following state-of-the-art sequence classification methods: MiSeRe \cite{egho_user_2017}, Sqn2Vec \cite{nguyen2019sqn2vec}, SGT \cite{ranjan_sequence_2021}, iBCM \cite{de_smedt_mining_2020}, SeqDT \cite{9411677}, SCIP \cite{zhou_pattern_2016}, and CR2N \cite{colleryneural}.

MiSeRe first mines sequential rules from the training data and then uses them as input for vector data classification algorithms. The $execution\_time$ and $max\_rule$ are fixed to 5 and 1024, respectively.

Sqn2Vec is a sequence embedding method that uses the mined patterns for training a vector representation. We set $min\_sup = 0.05$, $d = 128$ and $maxgap = 4$ for both its two variants, Sqn2VecSEP and Sqn2VecSIM.

SGT is a graph-based feature embedding method for converting sequences into feature vectors, for which we use the default parameters.

iBCM mines the sequential patterns based on the behavioral templates for sequence classification. We set $min\_sup = 0.1$, $no\_win =1$ and $reduce\_feature\_space = True$ for all data sets.

SeqDT integrates the correlated sequential pattern mining and the decision tree construction. We use the default parameters where $maxL=4$, $g=1$, $pru=ture$, $\epsilon = 1$, $minS=0.0$, $minN =2$, $maxD = 0$.

SCIP is a pattern-based sequence classification algorithm with four variants: SCII\_HAR, SCII\_MA, SCIS\_HAR, and SCIS\_MA. The following parameters are fixed in the experiments: $minint = 0.02$, $minsup=0.05$, $maxsize =3$, $conf = 0.5$ and $topk=11$.

CR2N is a 1D-convolutional neural network for discovering local or global patterns in sequential data while learning binary classification rules. For both local model and global model, $lr=0.1$ , $epochs =200$ , and regularisation $\lambda = 10^{-5}$.


\begin{figure*}[ht]
  \centering
  \includegraphics[scale = 0.35]{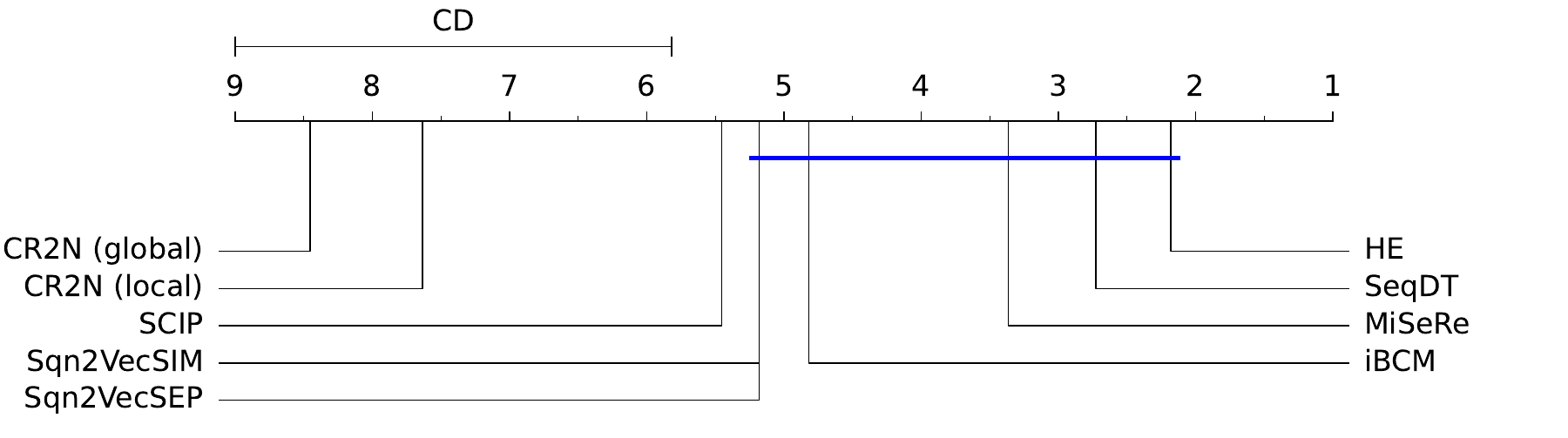}
  \caption{The comparison of Hamming Encoder (HE) against each competing method based on the Bonferroni-Dunn test. Our method is connected to one existing algorithm if their performance gap is not significantly different. SGT is not included because its classification results are not available on some data sets. }
  \label{fig:allCD}
\end{figure*}
\begin{figure*}[!ht]
    \begin{subfigure}{0.5\textwidth}
\centering
        \includegraphics[scale = 0.45]{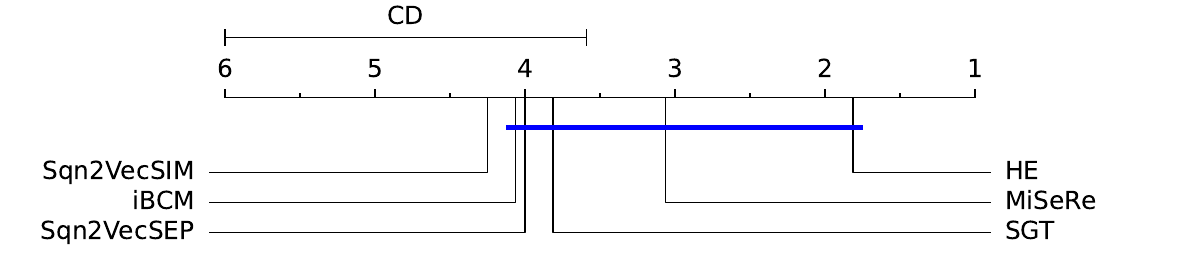}
        \caption{SVM}
        \label{fig:first}
    \end{subfigure}
    \hfill
    \begin{subfigure}{0.5\textwidth}
    \centering
        \includegraphics[scale = 0.45]{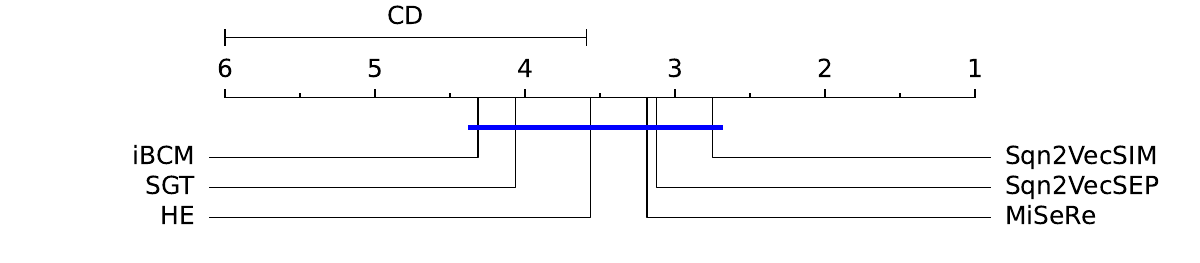}
        \caption{ NB}
        \label{fig:second}
    \end{subfigure}
    \\
    \begin{subfigure}{0.5\textwidth}
    \centering
        \includegraphics[scale = 0.45]{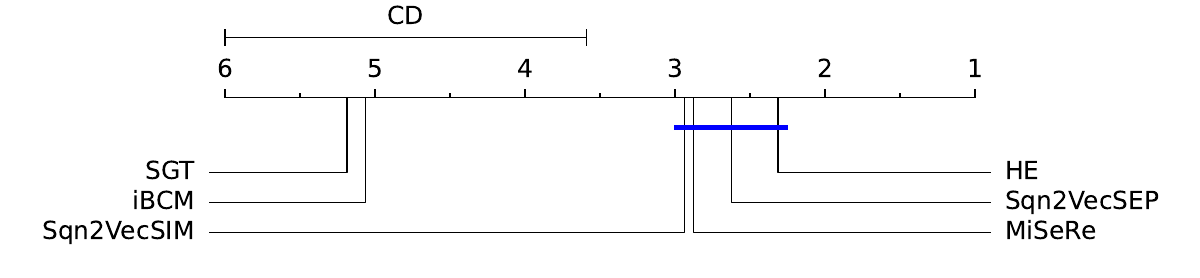} 
        \caption    {KNN}
        \label{fig:third}
    \end{subfigure}
    \hfill
    \begin{subfigure}{0.5\textwidth}
    \centering
        \includegraphics[scale = 0.43]{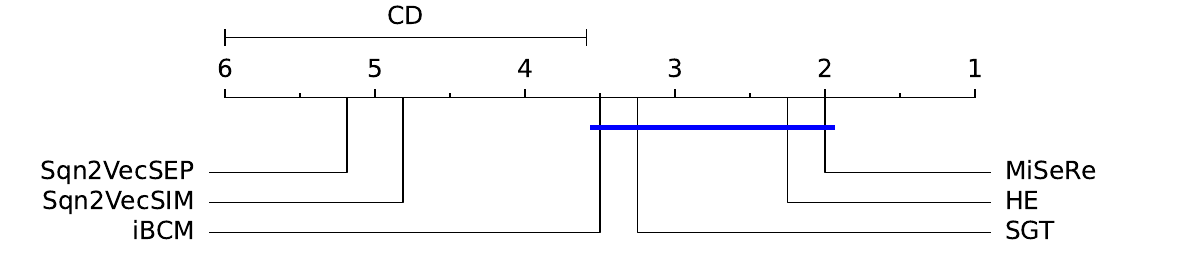}
        \caption    {DT}
        \label{fig:third}
    \end{subfigure}  
    \caption{The comparison of Hamming Encoder (HE) against each feature-based method coupled with four classifiers based on the Bonferroni-Dunn test (SGT is included).}
    \label{fig:nemenyi}
    \end{figure*}

\begin{figure*}[!ht]
    \begin{subfigure}{0.5\textwidth}
    \centering
        \includegraphics[scale = 0.45]{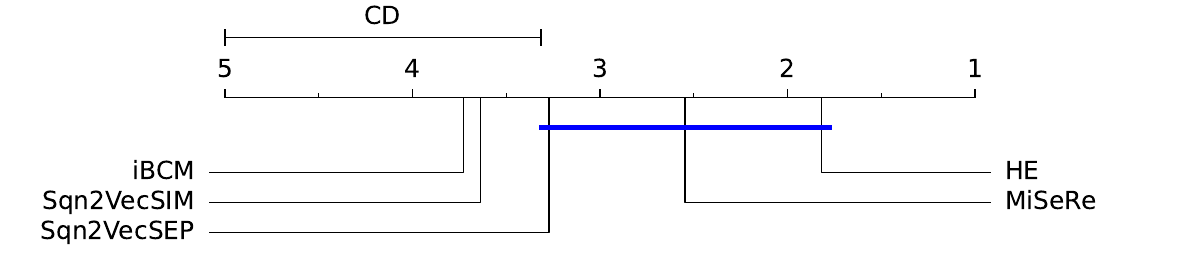}
        \caption{SVM}
        \label{fig:first}
    \end{subfigure}
    \hfill
    \begin{subfigure}{0.5\textwidth}
    \centering
        \includegraphics[scale = 0.45]{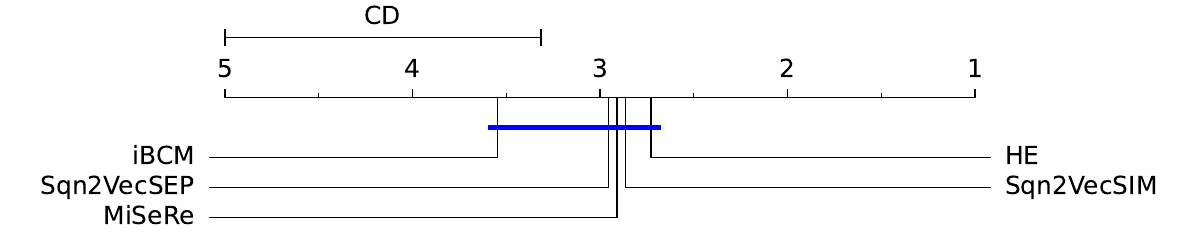}
        \caption{ NB}
        \label{fig:second}
    \end{subfigure}
    \\
    \begin{subfigure}{0.5\textwidth}
    \centering
        \includegraphics[scale = 0.45]{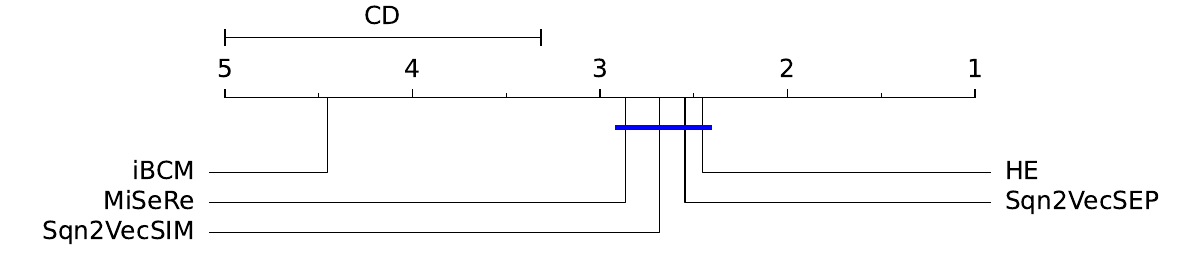} 
        \caption    {KNN}
        \label{fig:third}
    \end{subfigure}
    \hfill
    \begin{subfigure}{0.5\textwidth}
    \centering
        \includegraphics[scale = 0.45]{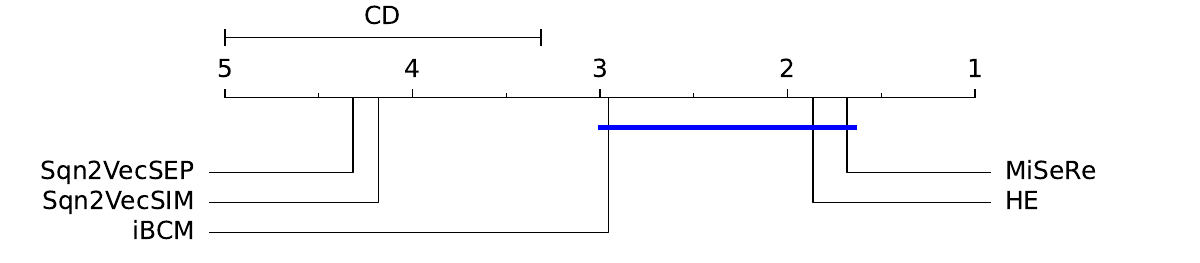}
        \caption    {DT}
        \label{fig:third}
    \end{subfigure}  
    \caption{The comparison of Hamming Encoder (HE) against each feature-based method coupled with four classifiers based on the Bonferroni-Dunn test (SGT is excluded).}
    \label{fig:withoutSGT}
    \end{figure*}
\subsection{Performance Comparison}

In our method, we set the initial number of $k$-mers to 1024 and the batch size to $64$. The length of $k$-mer is a user-specified parameter, and for our experiments, we roughly divided the candidate data sets into two classes based on the sequence length distribution. For data sets with a relatively large number of long sequences, we set $k=5$, and for those with a relatively large number of short sequences, we set $k=2$. The Adam optimizer is used with a learning rate of $0.0003$ and weight decay of $1e-5$. We select the best model based on the training set's loss and fix the number of iterations (epochs) to 100 for all data sets. We evaluate the discriminability of sequential patterns mined with different techniques using SVM, DT (Decision Tree), KNN (K-Nearest Neighbor), and NB (Naïve Bayes) in the scikit-learn library \cite{scikit-learn}.

\subsubsection{Comparison Results on Benchmark Data Sets}

The classification accuracies of different methods are listed in Table \ref{tab:results}. Some results of SGT are empty because its memory requirement exceeds the upper limit of our computer in the experiment. According to the experimental results in Table \ref{tab:results}, some important observations and comments are as follows.

Firstly, our method can achieve the highest classification accuracy on nearly half of the data sets. To evaluate the overall performance, we compute the average accuracy and rank, which are presented in the last two rows of Table \ref{tab:results}. Our method can achieve an average accuracy of 0.776, which outperforms the other methods. Moreover, our method's average rank of 2.182 is the best among all the methods employed in the comparison.

To further investigate the effectiveness of our method, we compute the average accuracy of the generated feature vectors with four different classifiers, as shown in Table \ref{tab:averageacc}. It shows that our method improves the performance of the SVM classifier compared to other methods. Similarly, our method outperforms other methods for the NB and KNN classifiers. However, for the DT classifier, our approach does not offer a clear performance gain.

To quantify whether our method really outperforms existing methods, the Bonferroni-Dunn test \cite{demvsar2006statistical} is employed to validate the null hypothesis that our method and each competing algorithm have the same performance in terms of classification accuracy.  The critical difference (CD) is $3.18$ when the significance level is $0.05$ and the number of data sets is $11$. If the difference between two methods in terms of the average rank is larger than CD, then the corresponding performance gap is statistically significant. The average rank of each method and the significance testing results are shown in Fig. 3. It can be observed that although Hamming Encoder cannot significantly beat all methods, it still leads over CR2N (global), CR2N (local), and SCIP. Meanwhile, this test is also applied for finding out whether the state-of-the-art feature-based methods and Hamming Encoder have the same performance on different classifiers. The result on $8$ data sets (SGT is included) is shown in Fig. \ref{fig:nemenyi} with $CD \simeq 2.41$ and the result on $11$ data sets (SGT is excluded) is shown in Fig. \ref{fig:withoutSGT} with $CD \simeq 1.68$. The significance testing results indicate that the Hamming Encoder does not show a significant advantage over existing methods when the classifier is NB. Furthermore, Hamming Encoder outperforms Sqn2VecSIM when the coupled classifier is SVM and leads over SGT and iBCM when coupled with KNN. Similarly, our method outperforms two Sqn2Vec variants when DT is used as the classifier.

\subsubsection{Loss over Training}
\begin{figure}[ht]
  \centering
  \includegraphics[scale = 0.37]{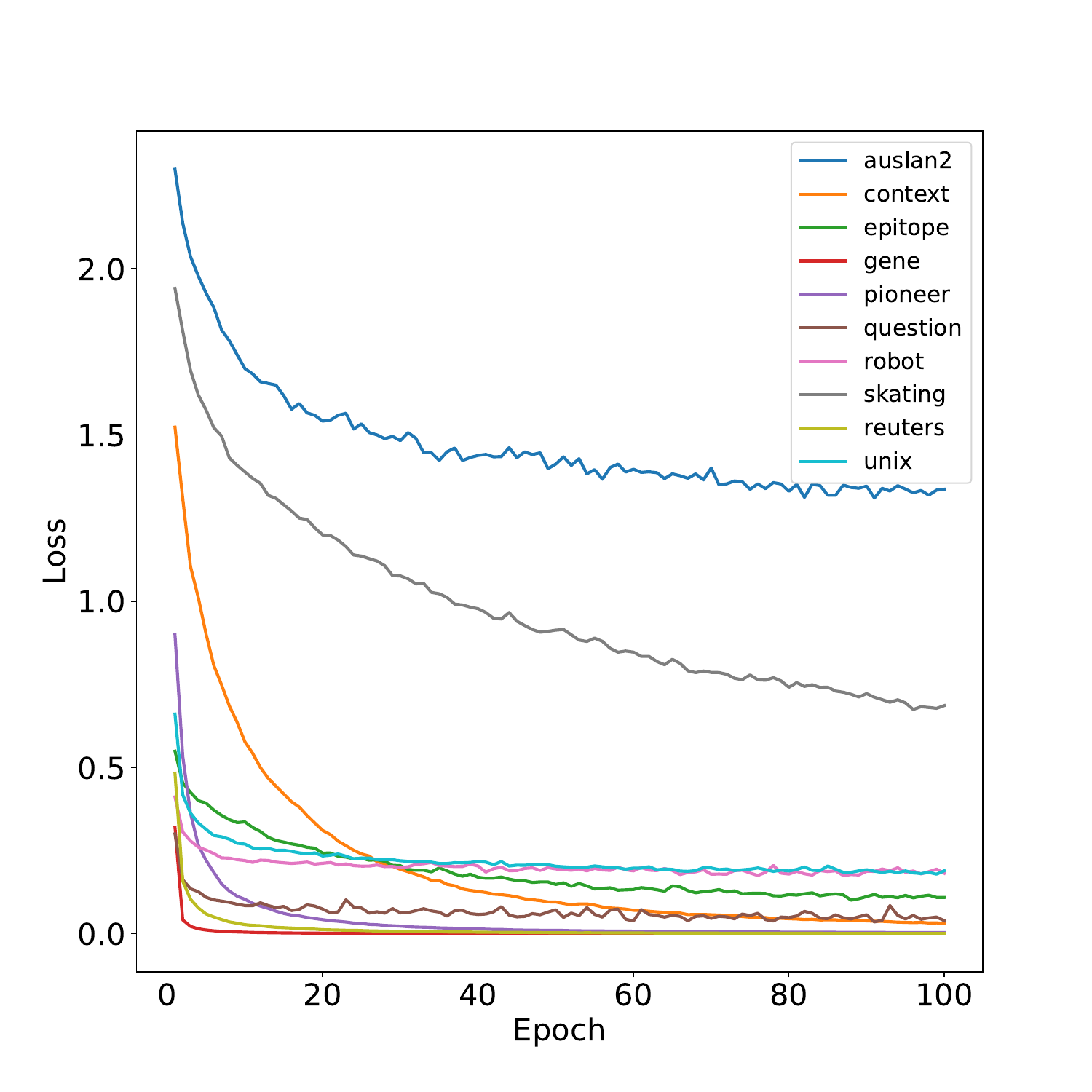}
  \caption{Training loss vs. Epochs for Hamming Encoder.}
  \label{fig:loss}
\end{figure}

In this section, we analyze the convergence behavior of our model by examining the trend of the loss function over the training epochs.

As can be seen from Fig. \ref{fig:loss}, the loss function shows a steady downward trend as the epoch increases. For most data sets, the loss tends to level off when it drops to around $40$. This indicates that the STE strategy performs well for this binarization method, and the proposed interpretable quantifier steadily converges during model training.

Previous experiments on classification accuracy show that this binarization strategy can indeed extract discriminative $k$-mer sets for sequence classification. Additionally, this interpretable quantifier may provide insights to understand the 1DCNN model. We have previously used SteHeaviside as a binarization quantizer, where the trained CNN weight matrix exhibits 0 in most positions and 1 in a few positions. In fact, this already provides a certain degree of interpretability since the weights can be considered as a coupling of several patterns. In other words, SteHeaviside may extract more than one pattern in one kernel, while it is difficult to analyze.

On the other hand, the Hamming Encoder binarization quantizer proposed in this paper adds constraints during training so that each kernel is directed to represent only one $k$-mer. It is experimentally shown that the Hamming Encoder converges well to yield a discriminative $k$-mer set.

\begin{figure}[!ht]
  \centering
  \includegraphics[scale = 0.37]{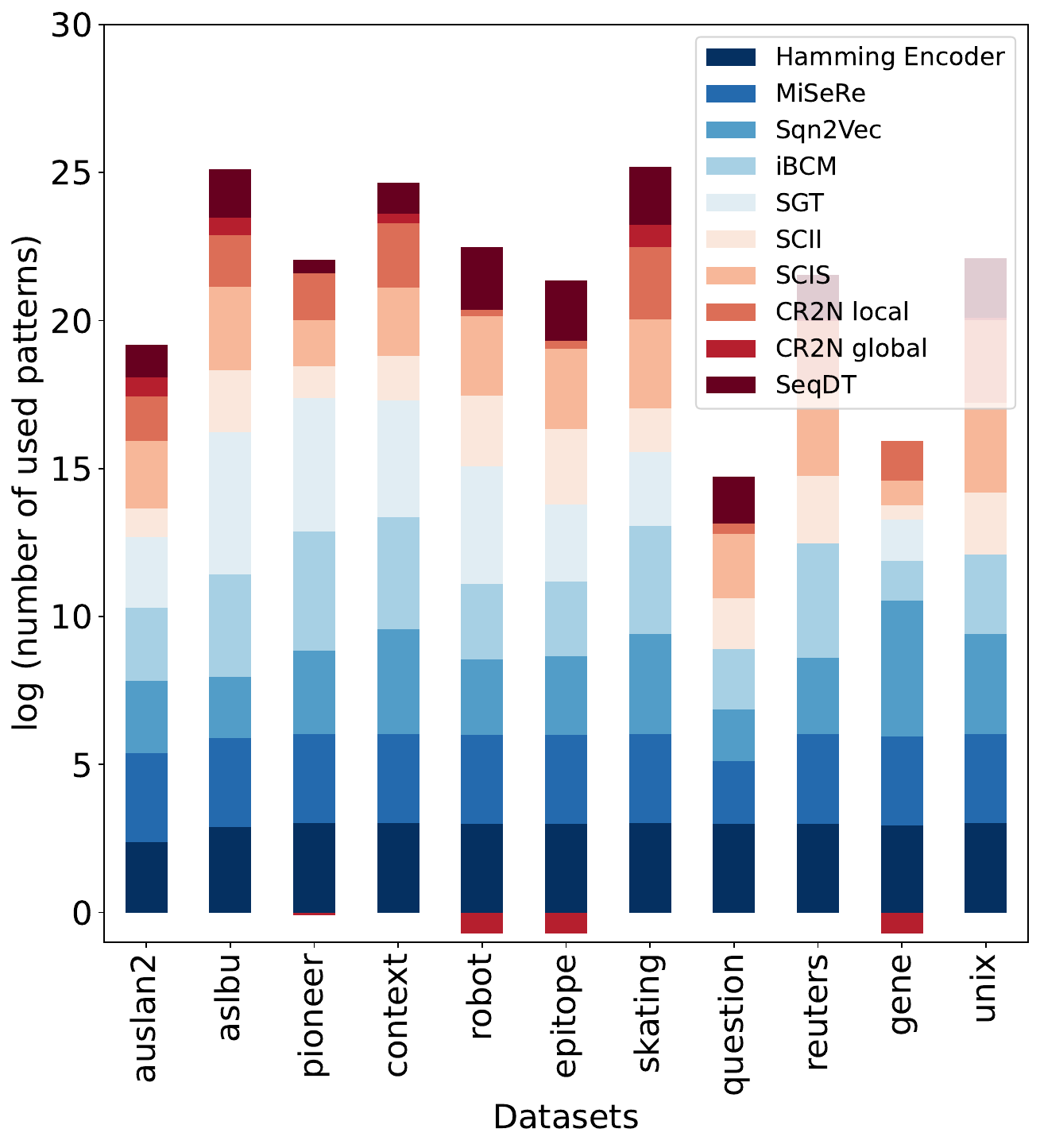}
  \caption{The comparison of different methods in terms of the number of features.}
  \label{fig:hist}
\end{figure}

\begin{figure}[ht]
  \centering
  \includegraphics[scale = 0.37]{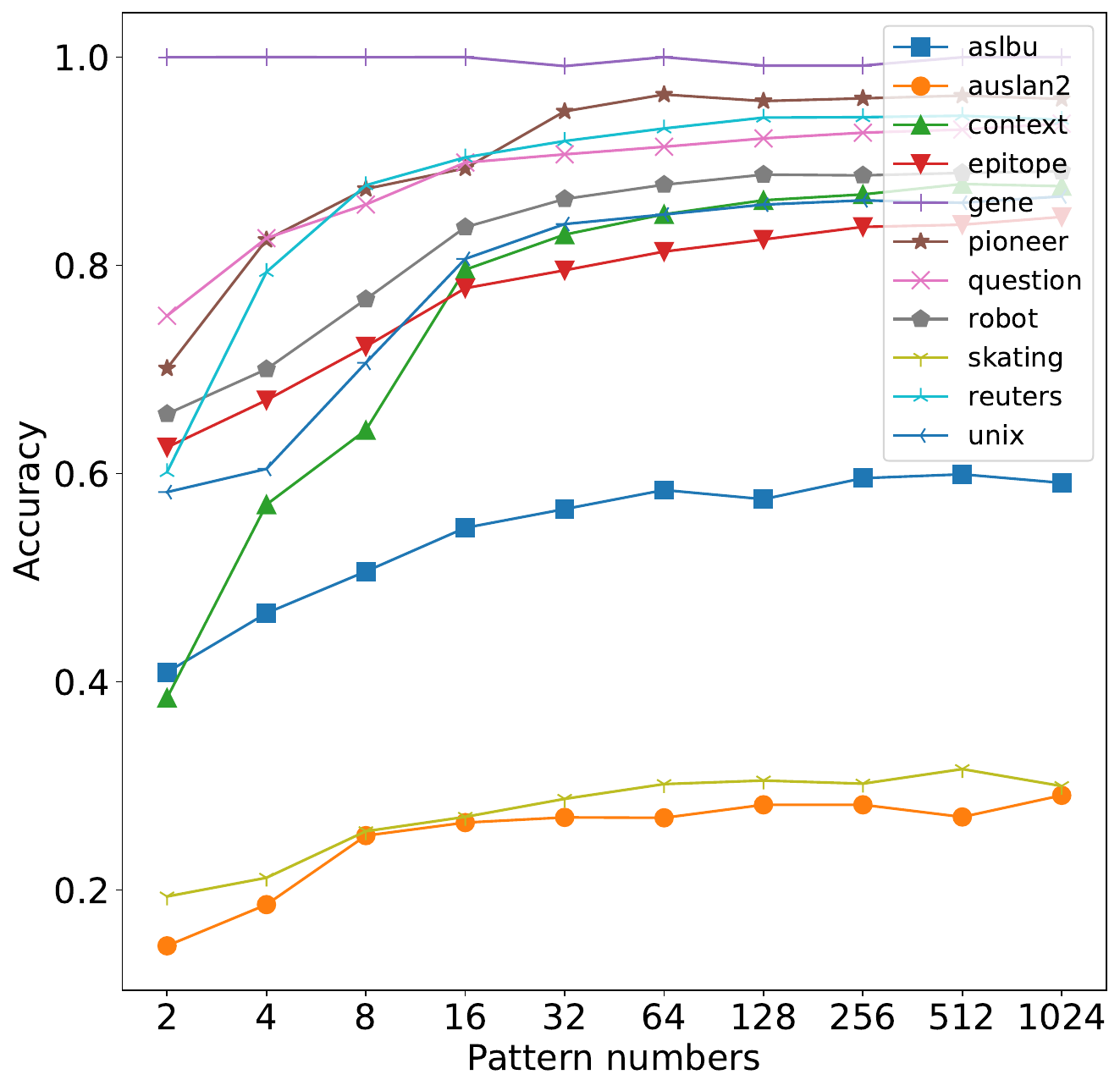}
  \caption{The fluctuation of accuracy per data set vs. the number of patterns mined.}
  \label{fig:patterns}
\end{figure}
\subsubsection{Number of Features}

Fig. \ref{fig:hist} presents the number of features used by different methods on benchmark data sets. It can be observed that SeqDT, CR2N (local), and CR2N (global) utilize a relatively small number of features across all data sets. This is primarily attributed to the integration of the pattern mining process with the classifier construction procedure. In particular, CR2N (local) is sensitive to certain local patterns and achieves remarkably high accuracy with a small number of patterns (rules). For example, on the ``robot" data set, CR2N (local) employs an average of only $1.6$ patterns and achieves $97.1\%$ classification accuracy. CR2N (global) reports very few patterns on most data sets, and in some cases, even zero rules are identified, such as in the ``gene" and ``epitope" data sets, which hinders its performance on these data sets.

On the other hand, iBCM and SGT sometimes output a large number of features for a given data set, resulting in unsatisfactory classification accuracy. For instance, SGT's classification accuracy is poor on the ``pioneer" and ``aslbu" data sets, while iBCM performs poorly on the ``reuters" and ``skating" data sets. This is primarily due to the unique characteristics of each data set, where iBCM employs a frequency-based screening strategy, and SGT's feature dimension can become high due to a large number of different items. In contrast, Hamming Encoder and MeSeRe adopt a Top-$k$ strategy and attain better classification accuracy with a controlled number of patterns. 

Fig. \ref{fig:patterns} presents the trend of the average classification accuracy of the Hamming Encoder as the number of patterns increases. It can be observed that the predictive performance generally improves as the number of patterns increases. However, this improvement stabilizes after the number of patterns used reaches a few hundred.

When examining the sequence information table alongside our proposed method's performance, it can be observed that the method generally performs well on data sets of longer sequences with a more uniform distribution of sequence lengths, such as ``question" and ``robot". However, it may not perform very well on data sets of shorter sequences with an uneven distribution of sequence lengths, such as ``auslan2" and ``unix". The reason for this is that our method employs a fixed-length $k$-mer, which may not be sufficient to capture enough information for shorter sequences. Increasing the $k$-mer length to accommodate shorter sequences would lead to many output $k$-mers being excluded from the data set, potentially resulting in overfitting during model training. As a result, our method may be less effective for data sets with short sequences and an uneven distribution of sequence lengths. 

\subsubsection{Running Time}
\begin{figure}[h]
  \centering
  \includegraphics[scale = 0.37]{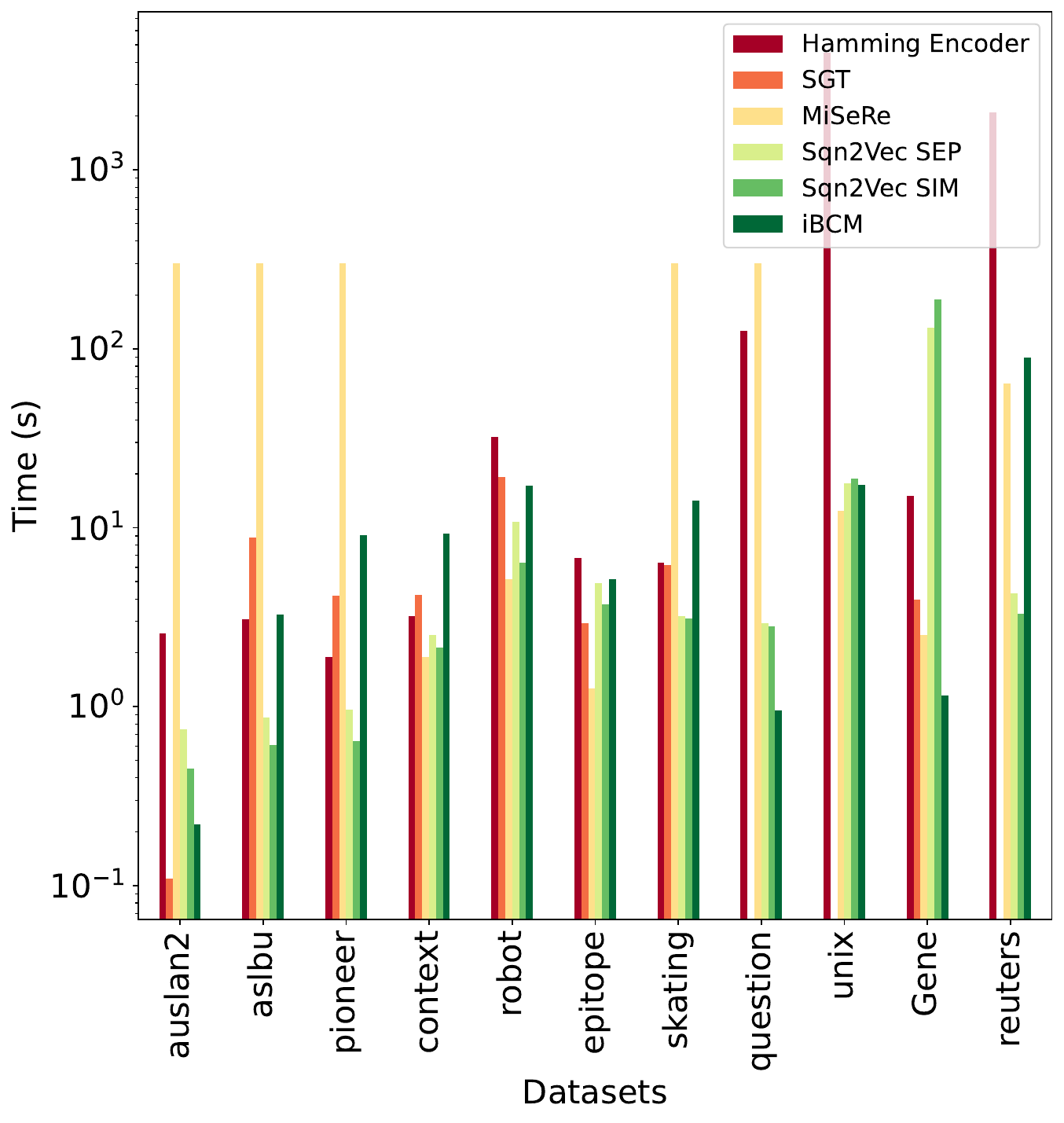}
  \caption{The comparison among feature-based methods in terms of the pattern mining time.}
  \label{fig:timetimetime}
\end{figure}
Fig. \ref{fig:timetimetime} provides a running time comparison among Hamming Encoder and other sequential pattern mining algorithms. The figure shows that our method is comparable to the other algorithms in most data sets. However, it underperforms on the ``unix" and ``reuters" data sets. This can be attributed to the fact that we utilize the 1DCNN structure for one-hot mapping and padding the matrix based on the maximum sequence length. As a result, the height of the input matrix is equal to the size of the item set, which means that the size of the convolutional kernel is bigger for data set like ``reuters". Besides, the computation of more convolution operations is necessary for data sets with larger maximum sequence lengths, such as ``unix". Therefore, the Hamming Encoder is relatively fast for mining a discriminative $k$-mer set in data sets with relatively uniform sequence length distribution and not too many items.

\begin{figure*}[!ht]
    
    \begin{subfigure}{0.5\textwidth}
    \centering
        \includegraphics[scale = 0.45]{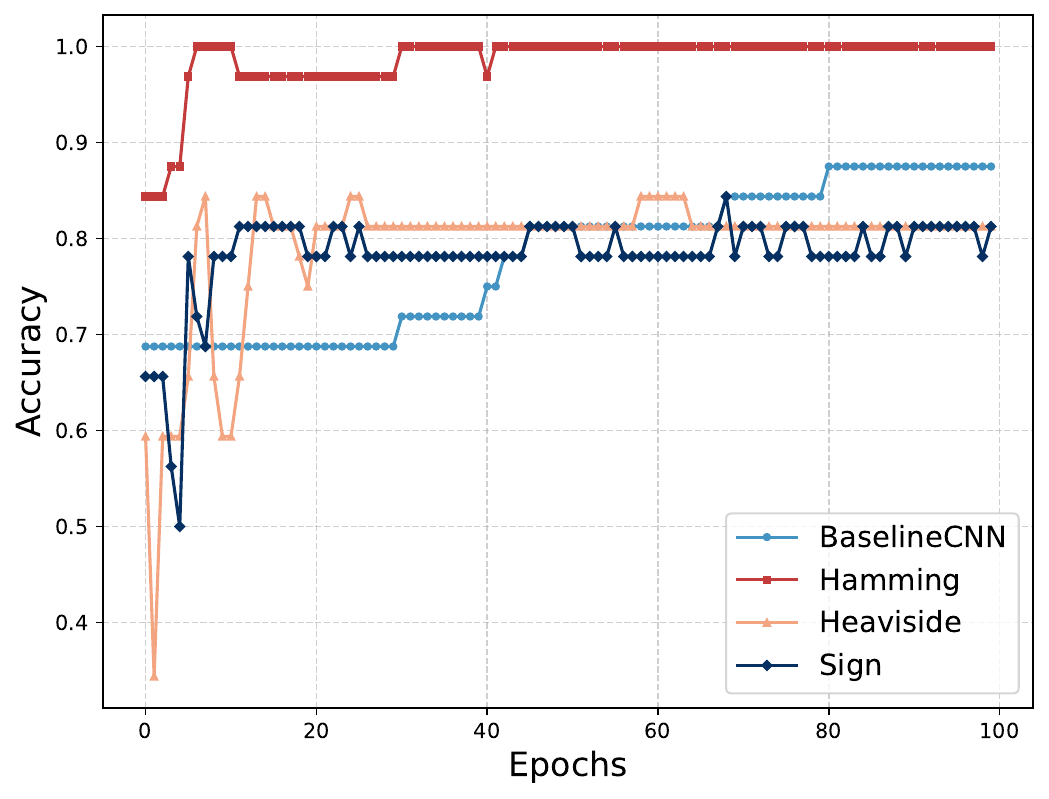} 
        \caption    {Pioneer}
        \label{fig:third}
    \end{subfigure}
    \hfill
    \begin{subfigure}{0.5\textwidth}
    \centering
        \includegraphics[scale = 0.45]{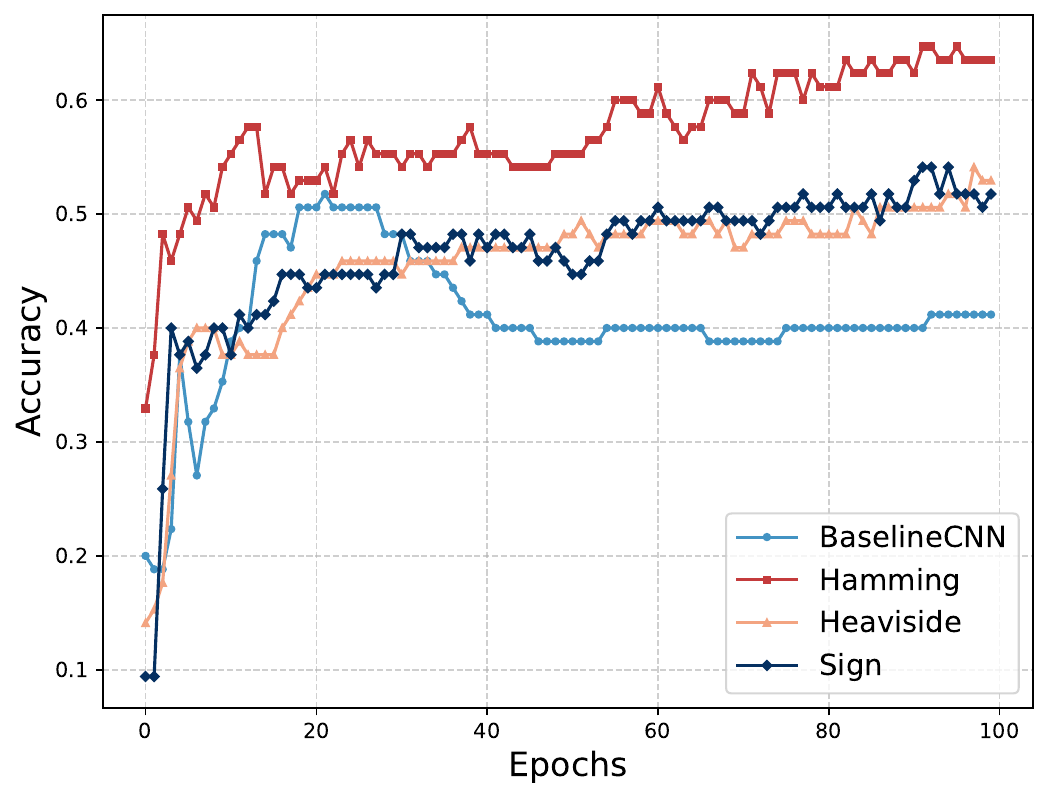}
        \caption    {Aslbu}
        \label{fig:third}
    \end{subfigure}  
    \caption{Performance evolution of Baseline CNN, Hamming, Heaviside, and Sign on ``pioneer'' and ``aslbu'' datasets with increasing epochs.}
   \label{twodatasets}
    \end{figure*}

\subsection{Advantages Compared to 1DCNN}
\begin{table}[]
\caption{Performance comparison of Hamming Encoder, Baseline CNN, Heaviside, Sign in terms of classification accuracy.}
\label{cnnbaseline}
\begin{tabular}{lllll}
\hline
Datasets & \textbf{Hamming}   & Baseline CNN & Heaviside & Sign     \\ \hline
aslbu    & \textbf{0.633}     & 0.446       & 0.524     & 0.552    \\
auslan2  & \textbf{0.277}     & 0.233       & \textbf{0.277}     & 0.275    \\
context  & 0.933     &\textbf{0.943}       & 0.941     & \textbf{0.943}    \\
epitope  & 0.896     & \textbf{0.945}       & 0.935     & \textbf{0.945}    \\
gene     & \textbf{1.000}     & \textbf{1.000}       & \textbf{1.000}     & \textbf{1.000}    \\
pioneer  & \textbf{0.985}     & 0.891       & 0.785     & 0.830    \\
question & 0.954     & \textbf{0.975}       & 0.950     & 0.948    \\
robot    & 0.904     & \textbf{0.974}       & 0.967     & 0.969    \\
skating  & \textbf{0.351}     & \textbf{0.351}       & 0.345     & 0.336    \\
reuters  & \textbf{0.979}     & 0.976       & 0.970     & 0.977    \\
unix     & 0.922     & \textbf{0.932}       & \textbf{0.932}     & \textbf{0.932}    \\ \hline
Average Acc &\textbf{0.803} & 0.788    & 0.784  & 0.790 \\ \hline
\end{tabular}
\end{table}
It is worth noticing first that our approach directly provides training strategies with interpretable weights. Hamming Encoder ensures that the weights can be translated into strings when constructing a classification model, which is particularly important for certain applications. For example, Hamming Encoder can efficiently construct pathogenic gene classifiers in large-scale gene databases while mining important gene fragments such as codons.

To investigate the comparative advantages of our approach in the context of sequence classification when contrasted with the widely used 1DCNN method, as well as prevalent binary methods: Heaviside $(0,1)$ and Sign $(-1,1)$,  we conducted an extensive series of experiments on eleven benchmark datasets. The parameters remained consistent with those of the Hamming Encoder.

Firstly, Table \ref{cnnbaseline} presents the classification accuracy of four end-to-end CNN methods. As evident from the table, our algorithm achieves the highest average accuracy of 0.803, which is superior to the baseline and two quantized CNNs.

Nevertheless, the binary strategy employed by the Hamming Encoder exhibits significant superiority, especially when on small-scale data sets. For instance, it achieves a notable performance increase of nearly 20\% compared to the baseline CNN and nearly 10\% to two quantized CNNs on ``aslbu''. In order to investigate the underlying reasons for these findings, we plotted the test accuracy versus the epoch curve in Fig \ref{twodatasets}. This analysis yielded several key insights. Notably, during the initial epochs, the Hamming Encoder already achieves higher accuracy on the test set compared to other CNNs. Furthermore, as training progresses, the Hamming Encoder consistently maintains higher accuracy than the baseline CNN and the alternative quantized methods. These observations can be attributed to the following reasons. For small-scale datasets, our binarization strategy actually limits the way of learning features, as it forces one kernel to focus on exactly one $k$-mer. Thus, it controls the model complexity by adding $ k$-mer-based regularization, which largely reduces overfitting and is especially beneficial for small-scale datasets.


\section{Conclusion and Future Work}
\label{conclusion}
In this paper, we propose a 1DCNN-based approach to mine discriminative $k$-mer sets for sequence classification. By formulating the mining of discriminative patterns as an optimization problem of an interpretable CNN encoder, our method effectively extracts a more discriminative $k$-mer set for classification with a gradient-based strategy. Our experimental results on a variety of benchmark data sets demonstrate that our mined $k$-mers achieve superior classification accuracy compared to state-of-the-art methods. 

Additionally, we introduce a novel similarity measure based on the Hamming distance to convert the mined $k$-mers into a feature vector, which ensures consistency with global max pooling operation in neural network architecture. It ensures interpretability during the forward computation of the neural network. Furthermore, our comparison results with other 1DCNN models in end-to-end training show that our model achieves the highest performance, demonstrating superior adaptability to overfitting.

However, our method still has several limitations. First, it can only report contiguous substrings, while in real-world scenarios, most sequence information contains gaps. Our current method cannot report substrings with gaps due to our binarization strategy. To address this issue, we need to propose better binarization strategies, such as selecting certain columns through statistical methods. Second, although our method can replace traditional 1DCNN neural networks, it still exhibits some performance degradation because some information is lost during the binarization process. Therefore, if we want to enhance the interpretability of 1DCNN neural networks using this binarization strategy in scenarios that require high accuracy, we need better gradient estimation methods. In future research, we will focus on solving the problem of mining substrings with gaps and proposing better gradient estimation methods.

\ifCLASSOPTIONcompsoc
  \section*{Acknowledgments}
\else
  \section*{Acknowledgment}
\fi

This work has been supported by the Natural Science Foundation of China under Grant No. 61972066.

\ifCLASSOPTIONcaptionsoff
  \newpage
\fi



%

\bibliographystyle{IEEEtran}
\bibliography{IEEEabrv,myrefs}

%

\begin{IEEEbiography}[{\includegraphics[width=1in,height=1.25in,clip,keepaspectratio]{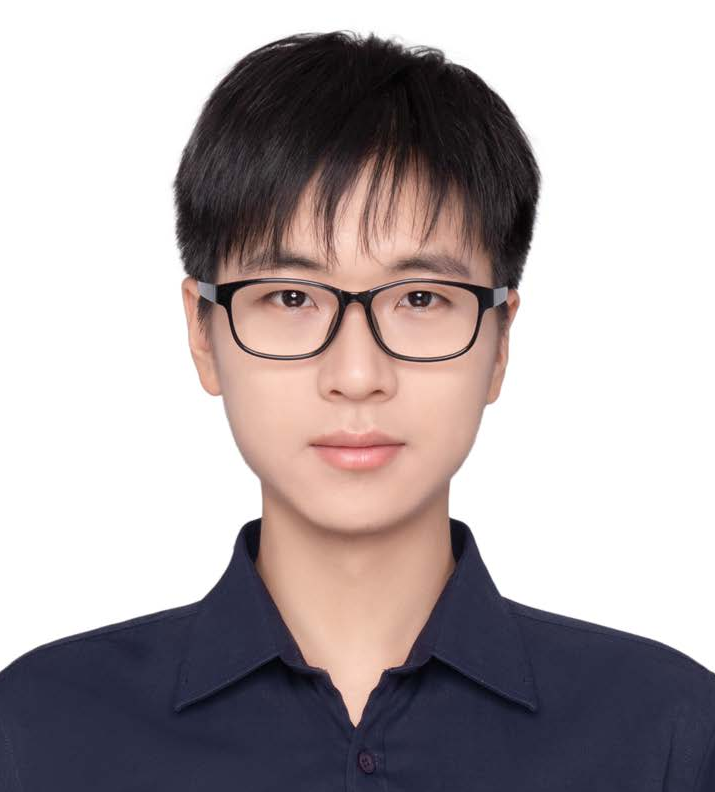}}]{Junjie Dong}
received the BS degrees in chemistry from Dalian University of Technology and University of Leicester in 2022, respectively. He is currently working toward the MS degree in the School of Software at Dalian University of Technology. His research interests include bioinformatics and data mining.
\end{IEEEbiography}

\begin{IEEEbiography}[{\includegraphics[width=1in,height=1.25in,clip,keepaspectratio]{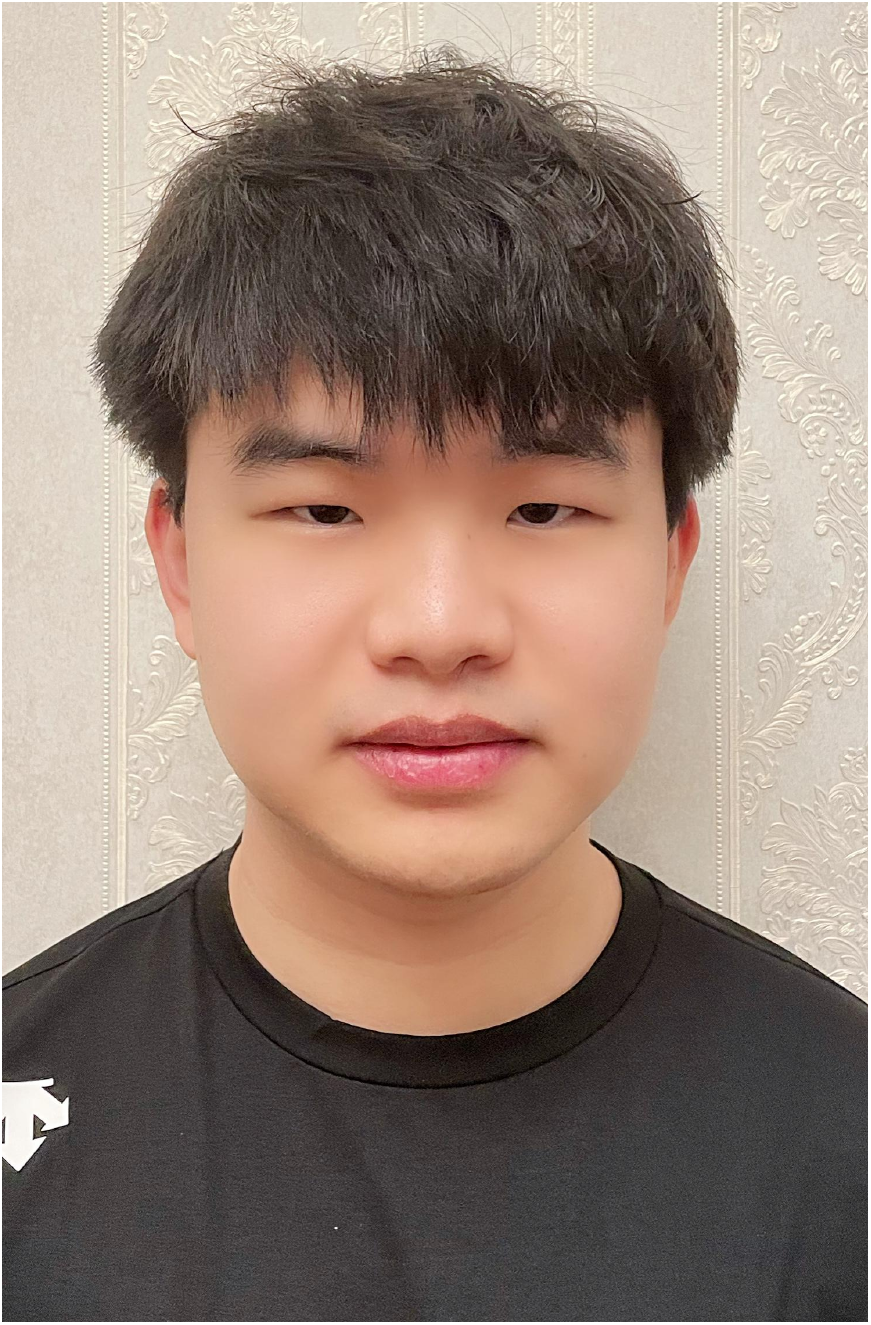}}]{Mudi Jiang}
received the BS degree in automation from Donghua University, China, in 2020. He is currently working toward the MS degree in the School of Software at Dalian University of Technology. His research interests include data mining and its applications.

\end{IEEEbiography}


\begin{IEEEbiography}[{\includegraphics[width=1in,height=1.25in,clip,keepaspectratio]{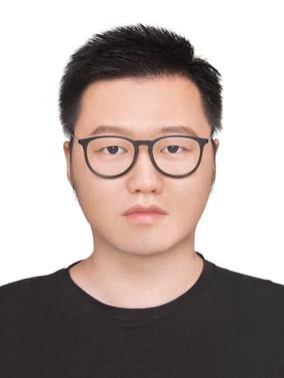}}]{Lianyu Hu}
received the MS degree in computer science from Ningbo University, China, in 2019. He is currently working toward the PhD degree in the School of Software at Dalian University of Technology. His current research interests include machine learning, cluster analysis and data mining.
\end{IEEEbiography}

\begin{IEEEbiography}[{\includegraphics[width=1in,height=1.25in,clip,keepaspectratio]{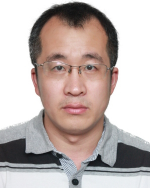}}]{Zengyou He}
received the BS, MS, and PhD degrees in computer science from Harbin Institute of Technology, China, in 2000, 2002, and 2006, respectively. He was a research associate in the Department of Electronic and Computer Engineering, Hong Kong University of Science and Technology from February 2007 to February 2010. He is currently a professor in the School of software, Dalian University of Technology. His research interests include data mining and bioinformatics.
\end{IEEEbiography}




\end{document}